\documentclass[lettersize,journal]{IEEEtran}
\usepackage{amsmath,amsfonts}
\usepackage{algorithmic}
\usepackage{algorithm}
\usepackage{array}
\usepackage[caption=false,font=normalsize,labelfont=sf,textfont=sf]{subfig}
\usepackage{textcomp}
\usepackage{stfloats}
\usepackage{url}
\usepackage{verbatim}
\usepackage{graphicx}
\usepackage{cite}
\usepackage{times}
\usepackage{epsfig}
\usepackage{graphicx}
\usepackage{amsmath}
\usepackage{amssymb}
\usepackage{bm}
\usepackage{nicefrac} 
\usepackage{microtype} 
\usepackage{blindtext}
\usepackage{tabulary,multirow,overpic}
\usepackage{booktabs}
\usepackage{makecell}  % xline
\usepackage{color}
\usepackage{balance}
\begin{document}

%%%%%%%%%%%%%%%%%%%%%%%%%%%%%%%%%%%%%%%% title %%%%%%%%%%%%%%%%%%%%%%%%%%%%%%%%%%%%%%%%
\title{AdaptVision: Dynamic Input Scaling in MLLMs for \\ Versatile Scene Understanding}
\author{
	\thanks{This work was supported by NSFC under Contract U20A20183 and 62021001. 
		It was also supported by GPU cluster built by MCC Lab of Information Science and Technology Institution, USTC, and the Supercomputing Center of the USTC.}
	
	Yonghui Wang,
    Wengang Zhou$^{\dagger}$,~\IEEEmembership{Senior Member,~IEEE},
	Hao Feng,
	and~Houqiang Li$^{\dagger}$,~\IEEEmembership{Fellow,~IEEE}
	\IEEEcompsocitemizethanks{\IEEEcompsocthanksitem Yonghui Wang, Wengang Zhou, Hao Feng, and Houqiang Li are with the CAS Key Laboratory of Technology in Geo-spatial Information Processing and Application System, Department of Electronic Engineering and Information Science, University of Science and Technology of China, Hefei, 230027, China.
	Wengang Zhou and Houqiang Li are also with Institute of Artificial Intelligence, Hefei Comprehensive National Science Center. 
	E-mail: \{wyh1998, haof\}@mail.ustc.edu.cn; \{zhwg, lihq\}@ustc.edu.cn.
	% \IEEEcompsocthanksitem  $^{\dagger}$Corresponding authors: Wengang Zhou and Houqiang Li.
}}

\markboth{IEEE TRANSACTIONS ON CIRCUITS AND SYSTEMS FOR VIDEO TECHNOLOGY,~Vol.~**, No.~**, July~2024}%
{Shell \MakeLowercase{\textit{et al.}}: A Sample Article Using IEEEtran.cls for IEEE Journals}
% after published
% \IEEEpubid{
% \begin{minipage}
% {\textwidth}\ \\[12pt] \centering Copyright~\copyright~2023 IEEE. Personal use of this material is permitted. 
% However, permission to use this material for any other purposes must \\ be obtained from the IEEE by sending an email to pubs-permissions@ieee.org.
% \end{minipage}
% }
\maketitle

%%%%%%%%%%%%%%%%%%%%%%%%%%%%%%%%%%%%%%%% abstract %%%%%%%%%%%%%%%%%%%%%%%%%%%%%%%%%%%%%%%%
\begin{abstract}
Over the past few years, the advancement of Multimodal Large Language Models (MLLMs) has captured the wide interest of researchers, leading to numerous innovations to enhance MLLMs' comprehension.
In this paper, we present AdaptVision, a multimodal large language model specifically designed to dynamically process input images at varying resolutions.
We hypothesize that the requisite number of visual tokens for the model is contingent upon both the resolution and content of the input image.
Generally, natural images with a lower information density can be effectively interpreted by the model using fewer visual tokens at reduced resolutions. 
In contrast, images containing textual content, such as documents with rich text, necessitate a higher number of visual tokens for accurate text interpretation due to their higher information density.
Building on this insight, we devise a dynamic image partitioning module that adjusts the number of visual tokens according to the size and aspect ratio of images.
This method mitigates distortion effects that arise from resizing images to a uniform resolution and dynamically optimizing the visual tokens input to the LLMs.
Our model is capable of processing images with resolutions up to $1008\times 1008$. 
Extensive experiments across various datasets demonstrate that our method achieves impressive performance in handling vision-language tasks in both natural and text-related scenes.
The source code and dataset are now publicly available at \url{https://github.com/harrytea/AdaptVision}.
\end{abstract}

\begin{IEEEkeywords}
Multimodal Large Language Model, Dynamic Image Resolution, Versatile Scene Understanding.
\end{IEEEkeywords}

%%%%%%%%%%%%%%%%%%%%%%%%%%%%%%%%%%%%%%%% Introduction %%%%%%%%%%%%%%%%%%%%%%%%%%%%%%%%%%%%%%%%%%%%
\section{Introduction}
\IEEEPARstart{S}{ince} the introduction of GPT-4~\cite{openai2023gpt4}, multimodal research has experienced unprecedented and rapid advancement. 
In the realm of Multimodal Large Language Models (MLLMs), the focus has been on integrating various modalities and augmenting model expertise through instruction tuning. 
In vision-language tasks, this technique not only improves the models' performance on specific tasks, but also highlights their remarkable zero-shot capabilities in visual-language comprehension and generation.
Consequently, it has led to the emergence of numerous influential works~\cite{alayrac2022flamingo,li2023blip,liu2024visual,zhu2023minigpt}.

\begin{figure}[tbp]
	\begin{center}
		\includegraphics[width=0.98\linewidth]{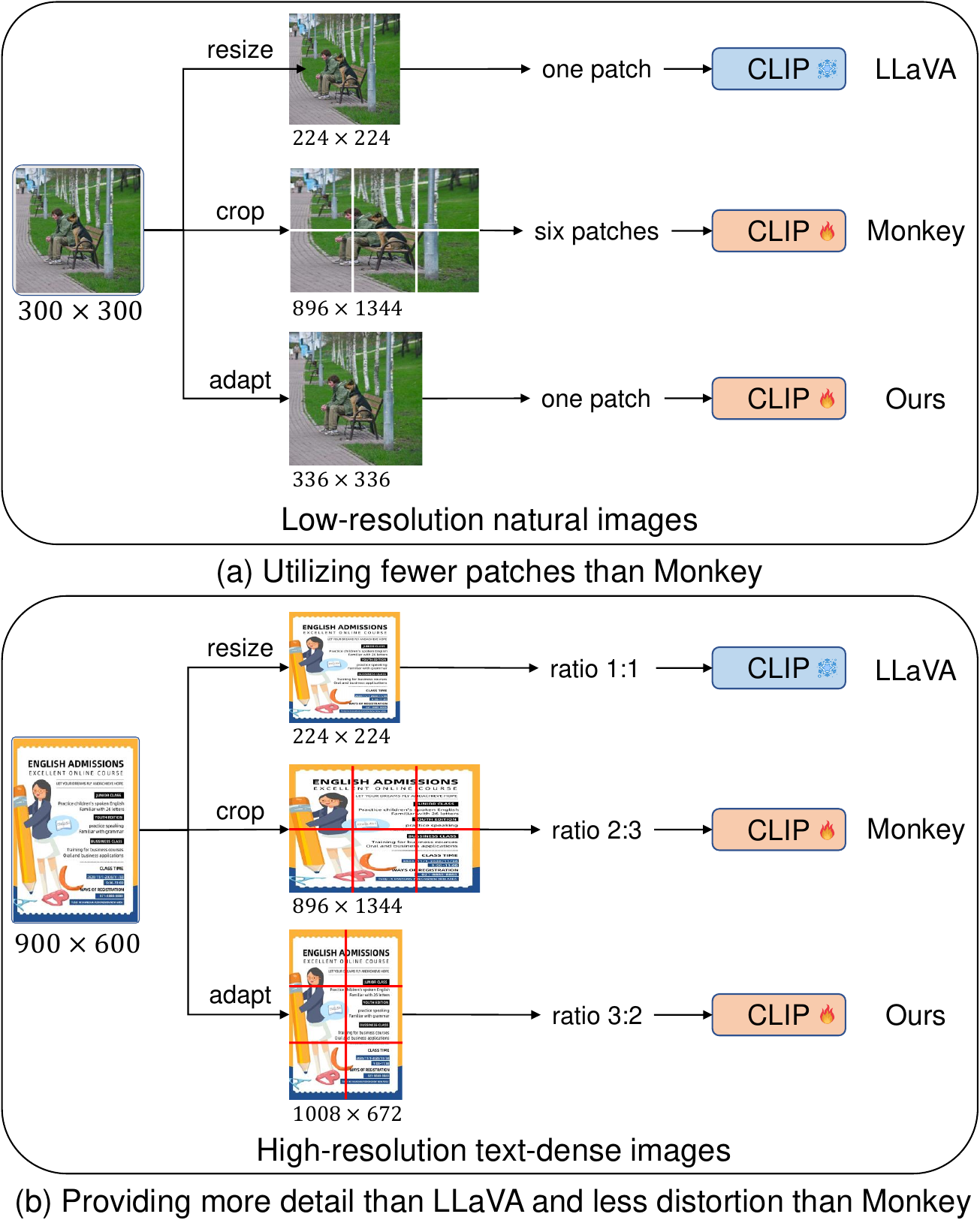}
	\end{center}
	\vspace{-0.1in}
	\caption{Comparisons of image processing with LLaVA~\cite{liu2024visual} and Monkey~\cite{li2023monkey}. Our method excels at optimizing patches by processing low-resolution natural images and adaptively adjusting the input for high-resolution text-dense images to mitigate the distortions of text within images from affecting overall comprehension.}
	\label{fig:intro}
\end{figure}

Recent studies, including LLaVA~\cite{liu2024visual} and MiniGPT-4~\cite{zhu2023minigpt}, have pioneered the exploration of multimodal large language models (MLLMs), allowing large language models (LLMs) to decode visual signals. 
However, their emphasis on training with natural images restricts their ability to parse textual details within the images.
To address this, a series of studies~\cite{zhang2023llavar,feng2023unidoc,wang2023towards,ye2023mplugdoc} have developed various strategies to generate instruction-following data for text-rich images, thus equipping models with textual interpretation skills such as optical character recognition (OCR).
Despite these advancements, it is evident that while the CLIP model~\cite{radford2021learning} is proficient in interpreting natural scenes, it struggles with the intricate details of granular text-related scenarios.
Consequently, several methods have been developed to improve perception of the resolution of the vision encoder, leading to significant progress~\cite{li2023monkey,feng2023docpedia,liu2023improved}.

Most current efforts to improve image resolution in MLLMs typically rely on a static resolution, resulting in a fixed number of visual tokens.
However, these methods might not be appropriate for images of varying types and sizes.
As shown in Figure~\ref{fig:intro}, we compare our technique with the image processing methods of LLaVA~\cite{liu2024visual} and Monkey~\cite{li2023monkey}. 
For low-resolution natural images with sparse information, the LLaVA method, as substantiated by quantitative results from Monkey's original paper, can effectively process these images, sometimes even surpassing Monkey.
This implies that high-resolution input introduces more visual tokens, some of which may be wasted, as MLLMs do not require an excess of visual tokens to comprehend natural images.
On the other hand, for high-resolution text-dense images, they require the analysis of detailed document images, which cannot be adequately handled by the low-resolution input of LLaVA. 
While Monkey can process high-resolution images, the fixed cropping strategy might distort the image, for instance, causing text deformation.
To address the above issue, we propose to dynamically adjust the number of visual tokens input into LLMs based on the size of the input image, allowing MLLMs to better comprehend the image while selecting the optimal resolution to reduce the distortion in the static CLIP-encoded images.

In this paper, we present AdaptVision, a method designed to adaptively process input images with varying resolutions, ensuring consistent model performance while regulating the number of visual tokens fed into the LLM.
Specifically, we develop a dynamic image partitioning module that is both simple in design and effective in performance.
The module initiates the process by forming a $3\times 3$ grid, with each cell's side length matching the input size of the vision encoder, and adjusts the input image to fit within the minimum bounding rectangle that fully encloses it.
The number of visual tokens fed into the LLM is determined based on the coverage of grid squares.
For images smaller in both dimensions than the vision encoder's input size, the partitioning step is skipped.
Moreover, inspired by~\cite{wang2023towards}, we enhance the model's ability to interpret texts within images by expanding the text-grounding instruction-following dataset to 100K samples.
We perform comprehensive evaluation in various tasks, including image captioning, general VQA, scene text-centric VQA, key information extraction, and document-related VQA, achieving significant results.

We summarize our contributions as follows:
\begin{itemize}
    \item We present AdaptVision, a method that dynamically adjusts the resolution based on the size and aspect ratio of images. This approach ensures consistent performance by using an appropriate number of visual tokens and reducing distortion.

    \item We enrich the text-grounding instructing-following data to 100K samples to enhance its interpretation ability on text-related tasks.

    \item We conduct extensive experiments across various datasets and tasks, showcasing the efficacy of our approach.
\end{itemize}
% \IEEEpubidadjcol

%%%%%%%%%%%%%%%%%%%%%%%%%%%%%%%%%%%%%%% Related work %%%%%%%%%%%%%%%%%%%%%%%%%%%%%%%%%%%%%%%%%%%%
\section{Related Work}
In this section, we begin with an overview of the development of multimodal large language models (MLLMs), followed by an analysis of studies proficient in interpreting text-rich images.
Finally, in response to the challenge posed by high text density in document images, we introduce the latest advancements in high-resolution document image processing techniques.

\subsection{MLLMs for Natural Scene Images}
Prior to the advent of GPT-4~\cite{openai2023gpt4}, the field of multimodal research has already made significant strides~\cite{zhu2019multi,liu2024hierarchical,guo2023dcmai,ma2023using,wu2023concept}, with numerous studies dedicated to creating modules to align the modalities of vision and text.
Specifically, Flamingo~\cite{alayrac2022flamingo} utilizes a perceiver resampler to selectively distill essential visual tokens, thereby achieving enhanced comprehension of visual signals in images through dense gated cross-attention blocks.
In the meantime, BLIP-2~\cite{li2023blip} leverages a pre-trained Q-former as a bridge to seamlessly integrate vision encoder outputs with large language models, improving the processing of visual information.
Despite these advancements, the substantial demand for data and computational resources has posed challenges for researchers, decelerating advances in multimodal large language model development.

The release of GPT-4 by OpenAI~\cite{openai2023gpt4} marks a pivotal milestone in the field of multimodal large language models (MLLMs).
To swiftly keep pace with GPT-4's advancements, Meta releases LLaMA~\cite{touvron2023llama}, a pre-trained large language model gaining widespread adoption in the academic community.
The model with 13B parameters surpasses the much larger GPT-3 model~\cite{brown2020language} with 175B parameters.
Subsequently, Vicuna~\cite{chiang2023vicuna}, the variant of LLaMA~\cite{touvron2023llama}, further enhances its capabilities.
Leveraging these advanced open-source large language models, a multitude of innovative research works have emerged, aiming to integrate visual signals with LLMs effectively.
LLaVA~\cite{liu2024visual} effectively transforms visual features encoded by CLIP~\cite{radford2021learning} into word embeddings through a simple linear layer and utilizes GPT-4 to refine the image-text pairs, leading to comprehensive instruction-following datasets for image-text alignment.
After finetuning the projector and the large language model, it achieves notable results. 
Meanwhile, MiniGPT-4~\cite{zhu2023minigpt} inherits the Q-former structure from BLIP-2~\cite{li2023blip} and introduces a linear mapping layer between image features and word embedding space.
Using GPT-4, it generates detailed image-text descriptions that are then subjected to manual verification to ensure data quality. 
With this high-quality dataset, the model employs a two-stage training strategy and significantly outperforms previous methods~\cite{li2023blip}.
Building on these foundational works, subsequent models such as LLaVA-1.5~\cite{liu2023improved} and MiniGPT-v2~\cite{chen2023minigpt}, along with other innovative works such as mPLUG-Owl~\cite{ye2023mplug}, InstructBLIP~\cite{dai2024instructblip}, and Shikra~\cite{chen2023shikra}, have further advanced the multimodal research field.

The above methods primarily focus on natural scenes and lack text Optical Character Recognition (OCR) capabilities for environments rich in text, leading to difficulties in handling text-related tasks.
This limitation is particularly evident in text-heavy scenes, where the performance of these models further declines. 
To address this issue, Multimodal Large Language Models (MLLMs) optimized specifically for text-rich scenes have been developed.

\subsection{MLLMs for Text-rich Images}
LLaVAR~\cite{zhang2023llavar} advances LLaVA's~\cite{liu2024visual} capabilities by constructing directive fine-tuning data using images with text. 
Initially, it utilizes existing OCR engines to gather a large corpus of text-rich image data, then employs GPT-4~\cite{openai2023gpt4} to generate 16K entries of text-related conversation data.
This newly collected data enables a subsequent round of fine-tuning on LLaVA, enhancing its performance on text-related benchmarks.
In parallel, mPLUG-DocOwl~\cite{ye2023mplugdoc} tackles the challenge of MLLMs' limited comprehension of diverse digital documents by introducing a unified multimodal framework for OCR-free document analysis. 
It curates a fine-tuning dataset encompassing an extensive array of vision-language understanding tasks, allowing the model to concentrate on fine-grained OCR features within images and perform well in language-only, general vision-and-language, and document understanding tasks. 
UniDoc~\cite{feng2023unidoc} stands out as the first to integrate text detection, recognition, spotting, and understanding into a single unified MLLM framework and finds that these tasks can mutually improve each other.
By crafting a large-scale fine-tuning dataset to bridge the gap between pre-training and fine-tuning stages, it applies text detection, recognition, and spotting tasks throughout both phases, with an emphasis on document comprehension during fine-tuning to enhance performance in text-heavy scenes.
TGDoc~\cite{wang2023towards} incorporates text-grounding into MLLMs, suggesting that guiding the model to identify the answer location via a bounding box during queries not only boosts interpretability but also reduces hallucinations.
By curating an instruction-following dataset including bounding boxes of text, they showcase the efficacy of their method.

The visual features of the aforementioned works are extracted by a frozen vision encoder, CLIP~\cite{radford2021learning}.
An alternative approach to handle fine-grained text details involves training a specialized transformer model, which mirrors the CLIP's~\cite{radford2021learning} architecture.
Vary~\cite{wei2023vary} observes that while the vision vocabulary derived from CLIP~\cite{radford2021learning} is adequate for many applications, it falls short in representing more complex tasks, such as document-level OCR. 
To overcome this limitation, Vary introduces a two-stage method to enrich the vision vocabulary.
Initially, it trains a custom vocabulary model via autoregression to identify fine-grained visual features.
Then, in a subsequent stage, this newly developed vocabulary is combined with the CLIP vision vocabulary and integrates the visual tokens into the LLM together, thus improving document-level comprehension.
Building on Vary, Vary-toy~\cite{wei2024small} further refines the representation of the vision vocabulary.
Specifically, it uniquely enhances the vocabulary by replacing negative samples of natural images with positive ones from object detection, optimizing visual information encoding.
The augmented vocabulary, paired with the smaller model Qwen-1.8B~\cite{bai2023qwenori}, can be deployed on a GTX1080Ti GPU and exhibits remarkable performance.

\begin{figure*}[tbp]
	\begin{center}
		\includegraphics[width=0.90\linewidth]{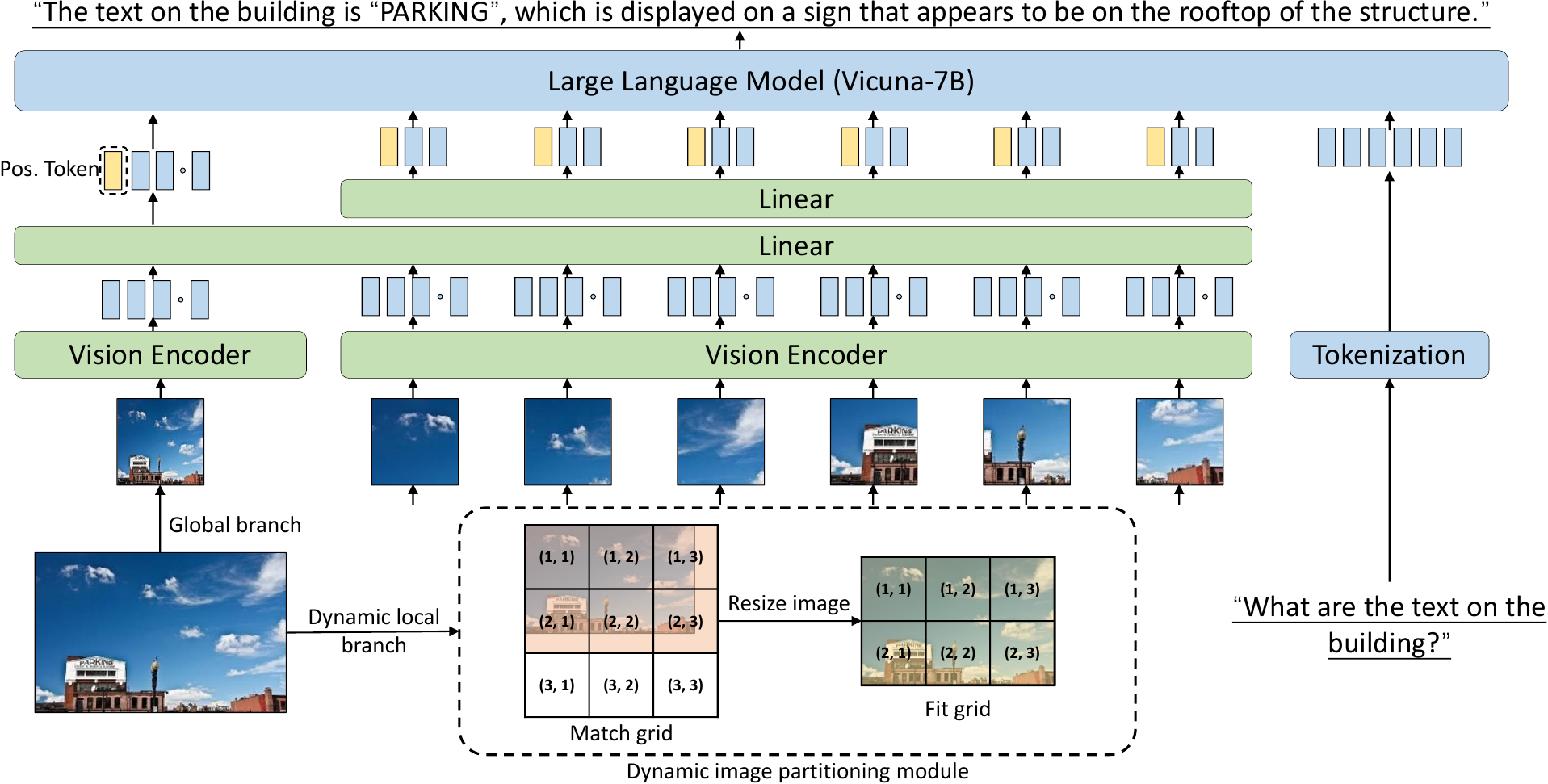}
	\end{center}
	\vspace{-0.1in}
	\caption{Overall architecture of AdaptVision. The process begins by splitting an image into two parts. The first is fed into the vision encoder, capturing the global information of the entire image. Meanwhile, the second undergoes adaptive segmentation via a dynamic image partition module, resulting in uniform patches that represent local features. Global features are directly projected into the word embedding space via a single linear layer. In contrast, local features undergo two layers of processing: the first aligns the dimensions with word embedding space, while the second performs dimensionality reduction, reducing the tokens to a quarter of their original count. In addition, learnable position tokens are prepended to both the global and local features to incorporate spatial context. Finally, all visual and text tokens are integrated into the LLM for further processing.}
	\label{fig:arch}
\end{figure*}

\subsection{MLLMs with High-resolution Input}
Constrained by CLIP's~\cite{radford2021learning} input resolution limits, Multimodal Large Language Models (MLLMs) typically process image inputs at resolutions such as $224\times 224$, $336\times 336$, and $448\times 448$, which restricts the depth of detailed scene understanding and high-resolution text recognition.
Monkey~\cite{li2023monkey} uniformly segments input images into patches, processing each with an independent CLIP~\cite{fang2023eva} while preserving the global features of the entire image. 
This method supports resolutions up to $1344\times 896$, demonstrating significant improvements across various benchmarks.
UReader~\cite{ye2023ureader} introduces a shape-adaptive cropping module that optimally prepares input images reasonably before feeding them to CLIP, allowing the frozen low-resolution vision encoder to handle high-resolution images.
DocPedia~\cite{feng2023docpedia}, an OCR-free document understanding model, further extends the input resolution to $2560\times 2560$.
This approach shifts visual image processing from pixel to the frequency domain, efficiently handling broader visual and textual content with fewer visual tokens.
TextMonkey~\cite{liu2024textmonkey} specializes in text-related tasks, applying shifted window attention to high-quality image processing for better efficiency.
They hypothesize that LLMs often process redundant visual tokens, and through selective filtering, they filter out unimportant tokens, retaining only the most informative ones.
This strategy improves model performance while markedly reducing the number of visual tokens.
These advances have shown exceptional performance in text and document-related tasks.

In this study, we propose AdaptVision, a model designed for a comprehensive understanding of various scenes, including natural and text-related scenes. 
AdaptVision innovatively adjusts the number of visual tokens to modulate the amount of visual information supplied to the Large Language Model (LLM), catering to both low-resolution natural scenes and high-resolution document images. 
Furthermore, through the compilation of existing instruction-following datasets, we enhance the model's proficiency in tasks such as image captioning, visual question answering, and key information extraction.
This integration endows the model with zero-shot capabilities, considerably expanding its applicability across diverse domains.

%%%%%%%%%%%%%%%%%%%%%%%%%%%%%%%%%%%%%%%%% Methods %%%%%%%%%%%%%%%%%%%%%%%%%%%%%%%%%%%%%%%%%%%%%%%
\section{Methods}
In this section, we begin by outlining the overall architecture of the model. 
Then, we detail the core input processing component, \emph{i.e.,} the dynamic image partitioning module. 
Finally, we discuss the tuning process, including the conversation format, the tuning tasks, and the datasets utilized for tuning.

\subsection{Architecture of AdaptVision}
The overall architecture of our method is illustrated in Figure~\ref{fig:arch}, which encompasses three main components: the vision encoder, the projector, and the Large Language Model (LLM). 
We use CLIP-ViT-L-336~\cite{radford2021learning} as our vision encoder, which incorporates a global branch to capture global information, along with a dynamic local branch for detailed insight.
The projector consists of two simple linear layers: one that projects image features into the word embedding space and another designed to minimize the number of visual tokens.
For our LLM, we employ Vicuna-7B~\cite{chiang2023vicuna}, which is used to interpret both image and text information and generates final outputs in an autoregressive manner.

Our method adaptively processes input images based on their aspect ratios and sizes, generating visual tokens that correspond to the dynamic image partitioning module.
This approach allows for precise control over visual tokens for various image types, enhancing comprehension on both global and local levels to improve efficiency.
The entire process is as follows:
Given an image $\bm{I}^{H\times W\times C}$, it is first passed to the global branch to capture the overarching information and extract global image features $\bm{F}_{g}^{576\times 1024}$.
Simultaneously, the dynamic local branch segments the image into uniform patches $\bm{P}_{i}^{336\times 336\times 3}$, where $i=0,1,2,\ldots$ represents the patch sequence.
Each patch $\bm{P}_{i}^{336\times 336\times 3}$ is then independently encoded to derive local image features $\bm{F}_{{l}_{i}}^{576\times 1024}$.
The visual tokens from both processing branches are projected into the word embedding space through a linear layer, producing $\bm{F}_{tg}^{576\times 4096}$ for global features and $\bm{F}_{{tl}_{i}}^{576\times 4096}$ for local features.
Subsequently, local features $\bm{F}_{{tl}_{i}}^{576\times 4096}$ are further compressed four times by an additional linear layer, resulting in condensed local features $\bm{F}_{{ctl}_{i}}^{144\times 4096}$.
Moreover, learnable position tokens are initialized for both global and local features, placed at the beginning of each feature.
These optimized visual tokens and tokenized word embeddings are concatenated and fed into the LLM to generate the final output.

% compact local features $\bm{F}{{rtl}{i}}^{144\times 4096}$. 
% These optimized visual tokens, combined with tokenized word embeddings, are concatenated and introduced into the LLM to produce the final output.

\begin{figure}[tbp]
	\begin{center}
		\includegraphics[width=0.98\linewidth]{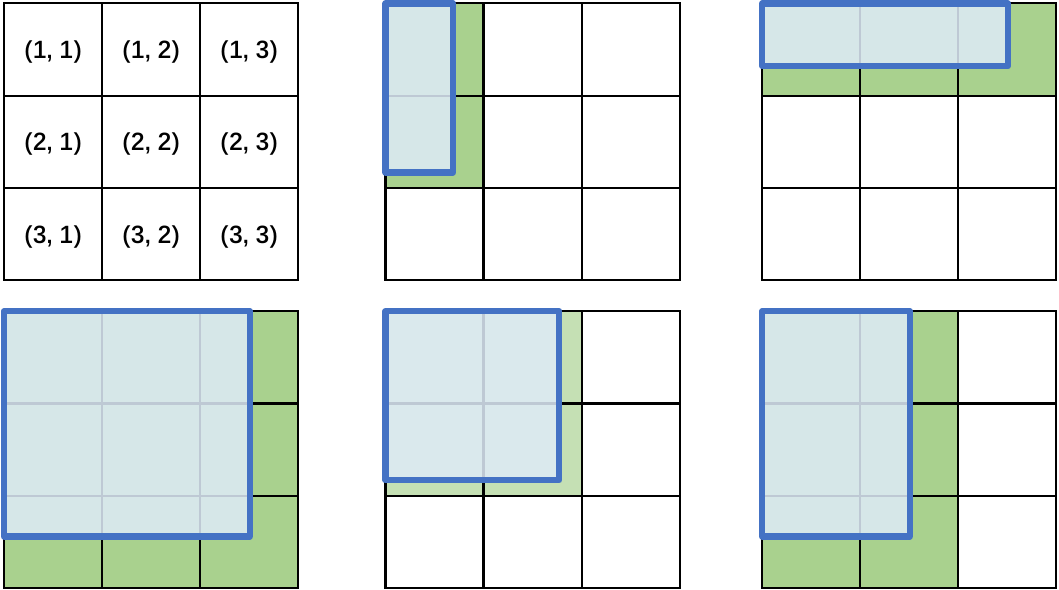}
	\end{center}
	\vspace{-0.1in}
	\caption{The principle of dynamic image partitioning module. We predefine a $3\times 3$ grid, where each grid cell is of the size of the input image. The subfigure in the upper left corner represents the positional tokens defined for each cell. In the remaining five subfigures, we exhibit some segment examples. The blue box outlines the image's original dimensions, while the green box indicates the image's resized dimensions, fitting the boundaries of the grid cell.}
	\label{fig:shape}
\end{figure}

\subsection{Dynamic Image Partitioning Module}
For MLLMs, processing inputs at higher resolutions allows the model to capture finer details within images.
However, when responding to user queries, it is not necessary to recognize all information within an image.
This is due to images comprising both crucial information necessary for addressing the problem and extraneous details that are not pertinent to the solution.

Building on our prior analysis, we propose a principle that dynamically adjusts the input resolution based on the size of the input image.
As shown in the upper left subfigure of Figure~\ref{fig:shape}, we define a grid $3\times 3$, where the size of each grid cell matches the input dimension of the vision encoder. 
We require the resized image to fit the boundaries of the grid cell.
As shown in the remaining five subfigures of Figure~\ref{fig:shape} , we exhibit some segment examples. 
The original dimensions of the image are represented by a blue box, while the corresponding green box indicates the image's resized dimensions, fitting the boundaries of the grid cell.
This method enables the model to eventually simulate the processing of images with a maximum resolution of $1008\times 1008$.
It should be noted that for input images smaller than $336\times 336$, the dynamic branch is also performed.
On the other hand, images larger than $1008\times 1008$ will be resized to a maximum resolution of $1008\times 1008$.
The dynamic image partitioning module scales flexibly with the input image dimensions, supporting aspect ratios from 1:1 to 3:3, including intermediate ratios like 1:2, 1: 3 and 2:1.
The method accommodates pixel dimensions from $336\times 336$ to $1008\times 1008$, with increments of 336 pixels delineating each scaling step.
We believe that the versatility covers the needs of most practical applications.

Within the defined grid, each grid cell corresponds to an individual image patch.
To mitigate the loss of patch location information due to variability in input image sizes, we introduce the position tokens for each grid cell.
As illustrated in the upper left corner of Figure~\ref{fig:shape}, we predefined nine grids, segmented from (1, 1) to (3, 3), and designated the previous global image patch as (0, 0).
Thus, we initialize a total of ten learnable position tokens, which are concatenated at the beginning of the corresponding image features.
This helps the model in accurately determining the location of each image patch, improving its ability to grasp the overall context.

\subsection{Instruction Tuning}
In this section, we introduce the conversation format of instruction tuning for pre-training and fine-tuning stages, followed by a detailed description of the tasks chosen for the two stages and the datasets utilized.

\textbf{Conversation format.}
We offer the template for the general training tasks as follows:

% \textit{\textless System Message\textgreater USER: \textless image\textgreater \textless Image Embedding\textgreater \textless /image\textgreater \\Question + Instruction ASSISTANT: \textless Model's Output\textgreater}
\textit{\textless System Message\textgreater}

\textit{USER: \textless image\textgreater \textless ImageEmbedding\textgreater \textless /image\textgreater Question + Instruction ASSISTANT: \textless Model Output\textgreater}

\noindent where the system message is ``A chat between a curious user and an artificial intelligence assistant. The assistant gives helpful, detailed, and polite responses to user questions''.
``Question'' refers to the queries asked by users, including image captioning, visual question answering, etc.
``Instruction'' denotes the directives given by the user.
For the text-grounding VQA, we instruct the model with ``Please provide the supporting text and its bounding box.''
For other queries, we refer to Monkey~\cite{li2023monkey}, employing the instruction ``Answer:'' to prompt the model to output the answer directly.

\begin{table}
    \centering
    \setlength\tabcolsep{1pt}
    \small
    \caption{Instruction-following data we used for pre-training.}
    \resizebox{0.98\linewidth}{!}{%
        \begin{tabular}{llc}
        \Xhline{2.5\arrayrulewidth}                                                  
        Task & Datasest & Conversation \\
        \hline
        Image Caption & CC3M & 595K \\
        \hline
        Text Detection & LAION-5B & 422K \\
        \hline
        Text Detection, Recognition, Spotting & PPT & 99K \\
        \Xhline{2.5\arrayrulewidth}
    \end{tabular}}
    \label{tab:pretrain}
\end{table}

\textbf{Tuning tasks and datasets.}
Our objective is to apply our method across a variety of scenarios. 
The model we have developed is capable of identifying different objects in a scene and recognizing textual information embedded within it. 
Similarly to LLaVA~\cite{liu2024visual}, we also employ a two-stage training strategy.
The pre-training stage involves training two projectors to align images and text, while the fine-tuning stage focuses on optimizing the entire pipeline to bolster the model's zero-shot learning capabilities.
Specifically, for the pre-training stage, we engage in tasks including image captioning, text detection, recognition, and spotting.
All the datasets are detailed in Table~\ref{tab:pretrain}, comprising 595K samples from LLaVA~\cite{liu2024visual}, 422K samples from LLaVAR~\cite{zhang2023llavar}, and 99K PPT samples from TGDoc~\cite{wang2023towards}.
During the fine-tuning stage, our aim is to adapt the model to a variety of tasks by leveraging instruction-following datasets with diverse instructions, thereby achieving zero-shot generalization capabilities.
We perform tasks including image captioning, general VQA, text-centric scene VQA, key information extraction, document-oriented VQA, and text-grounding VQA.
The quantities of datasets employed are listed in Table~\ref{tab:finetune}, incorporating the dataset from Monkey~\cite{li2023monkey}, along with 158K COCO samples from LLaVA~\cite{liu2024visual} and 16K LAION-5B samples from LLaVAR~\cite{zhang2023llavar}. 
Furthermore, we expand the text-grounding VQA data in TGDoc~\cite{wang2023towards} and name Text-grounding VQA to 100K samples to further enhance the model's ability in text processing.

\begin{table}
    \centering
    \setlength\tabcolsep{1pt}
    \small
    \caption{Instruction-following data we used for fine-tuning.}
    \resizebox{0.98\linewidth}{!}{%
        \begin{tabular}{llc}
        \Xhline{2.5\arrayrulewidth}                                                  
        Task & Dataset & Conversation \\
        \hline
        \multirow{4}{*}{Image Caption} & COCO Image~\cite{liu2024visual} & 158K \\
                                    & Detailed Caption~\cite{li2023monkey} & 213K \\
                                    & COCO Caption~\cite{karpathy2015deep} & 82K \\
                                    & TextCaps~\cite{sidorov2020textcaps} & 109K \\
        \hline
        \multirow{5}{*}{General VQA} & VQAv2~\cite{goyal2017making} & 100K \\
                                    & OKVQA~\cite{marino2019ok} & 18K \\
                                    & GQA~\cite{hudson2019gqa} & 150K \\
                                    & ScienceQA~\cite{lu2022learn} & 18K \\
                                    & VizWiz~\cite{gurari2018vizwiz} & 20K \\
        \hline
        \multirow{4}{*}{\makecell[c]{Scene Text-centric \\ VQA}} & LAION-5B~\cite{schuhmann2022laion} & 16K \\
                                    & TextVQA~\cite{singh2019towards} & 34K \\
                                    & OCRVQA~\cite{mishra2019ocr} & 250K \\
                                    & AI2D~\cite{kembhavi2016diagram} & 24K \\
        \hline
        \multirow{8}{*}{Doc-oriented VQA} & DocVQA~\cite{mathew2021docvqa} & 118K \\
                                    & ChartQA~\cite{masry2022chartqa} & 84K \\
                                    & InfoVQA~\cite{mathew2022infographicvqa} & 47K \\
                                    & DeepForm~\cite{svetlichnaya2020deepform} & 7K \\
                                    & KLC~\cite{stanislawek2021kleister} & 27K \\
                                    & WTQ~\cite{pasupat2015compositional} & 28K \\
                                    & TabFast~\cite{chen2019tabfact} & 91K \\
                                    & VisualMRC~\cite{tanaka2021visualmrc} & 21K \\
        \hline
        Text-grounding VQA          & TextGroundVQA & 100K \\
            %			\hline
            \Xhline{2.5\arrayrulewidth}
    \end{tabular}}
    \label{tab:finetune}
\end{table}

\begin{table*}[t]
	\centering
	\setlength\tabcolsep{1pt}
	\small
	\caption{Quantitative comparison with previous multimodal large language models (MLLMs) across a variety of benchmarks and tasks, including image caption, general VQA, scene text-centric VQA, and key information extraction. The best and the second results are highlighted in \textbf{bold} and \underline{underlined}, respectively.}
	\resizebox{0.98\linewidth}{!}{%
		\begin{tabular}{lccccccccccccccc}
			\Xhline{2.5\arrayrulewidth}                                                  
			\multirow{3}{*}{Method} & \multirow{3}{*}{LLM} & \multicolumn{2}{c}{Image Caption} & \multicolumn{5}{c}{General VQA}  & \multicolumn{4}{c}{Scene Text-centric VQA} & \multicolumn{3}{c}{Key Information Extraction}  \\ 
			\cmidrule(lr){3-4}  \cmidrule(lr){5-9} \cmidrule(lr){10-13} \cmidrule(lr){14-16} 
			& & Flickr30K & TextCaps & VQAv2 & OKVQA & GQA & ScienceQA  & VizWiz & STVQA & OCRVQA & AI2D & TextVQA & FUNSD & SROIE & POIE \\
			%  & 5000  & 5000   & 5000    & 5349 & 2801 & 1250 & 588   & 2503  & 6321 & \\
			\hline
            BLIP2-OPT-6.7B~\cite{li2023blip} & OPT-6.7B & 61.90 & 54.22 & - & - & 38.10 & - & 33.75 & 13.36 & 10.58 & 42.55 & 21.18 & 0.00 & 0.00 & 0.02 \\
            InstructBLIP~\cite{dai2024instructblip} & Vicuna-7B & 82.06 & 78.18 & 75.62 & 56.08 & 49.48 & - & 32.86 & 28.30 & 51.00 & 34.29 & 39.82 & 1.02 & 0.59 & 2.14 \\
            mPLUG-Owl~\cite{ye2023mplug} & LLaMA-7B & - & - & - & - & 29.64 & - & 9.04 & 29.26 & 28.62 & 62.18 & 40.28 & 1.02 & 0.64 & 3.26 \\
            MiniGPT-4~\cite{zhu2023minigpt} & Vicuna-7B & - & - & - & 33.00 & 19.21 & - & 1.16 & 14.02 & 11.52 & 62.42 & 18.72 & 1.19 & 0.04 & 1.31 \\
            LLaVAR~\cite{zhang2023llavar} & Vicuna-7B & - & - & - & - & - & - & - & 30.36 & 29.38 & - & 39.40 & 1.02 & 1.36 & 6.48 \\
            UniDoc~\cite{feng2023unidoc} & Vicuna-7B & - & - & - & - & - & - & - & 30.78 & 34.50 & - & 40.72 & 1.19 & 1.40 & 3.92  \\
            BLIVA~\cite{hu2023bliva} & Vicuna-7B & - & - & - & - & - & - & - & 32.33 & \underline{64.04} & - & 44.02 & 2.04 & 0.72 & 3.70 \\
            LLaVA-1.5~\cite{liu2023improved} & Vicuna-7B & 52.98 & 67.02 & 78.50 & - & \underline{61.70} & 63.26 & 50.00 & 42.72 & 62.80 & 43.81 & 53.98 & 2.04 & 2.96 & 5.60 \\
            mPLUG-Owl2~\cite{ye2023mplug2} & LLaMA2-7B & 81.12 & - & 79.10 & 57.70 & 56.10 & 68.70 & 54.08 & 48.02 & 63.54 & 55.21 & 60.80 & 2.21 & 2.42 & 7.61 \\
            DocPedia~\cite{feng2023docpedia} & Vicuna-7B & - & - & - & - & - & - & - & 45.54 & 57.20 & - & 60.18 & \textbf{29.86} & 21.44 & \textbf{39.94} \\
            Qwen-VL~\cite{bai2023qwen} & Qwen-7B & 80.79 & 62.43 & 79.50 & 58.42 & 60.34 & 67.10 & 38.36 & 48.13 & 63.66 & 59.55 & 63.80 & 3.40 & 28.32 & 2.97  \\
            Monkey~\cite{li2023monkey} & Qwen-7B & \underline{82.18} & \textbf{92.75} & \underline{80.15} & \textbf{61.48} & 60.80 & \underline{69.40} & \underline{61.10} & \textbf{54.88} & 63.45 & \underline{62.66} & \underline{63.86} & \underline{23.43} & \textbf{41.00} & 17.26 \\
            \textbf{AdaptVision} & Vicuna-7B & \textbf{83.32} & \underline{86.81} & \textbf{81.50} & \underline{58.87} & \textbf{62.37} & \textbf{71.39} & \textbf{68.38} & \underline{49.26} & \textbf{64.36} & \textbf{66.65} & \textbf{66.26} & 22.28 & \underline{38.50} & \underline{36.67} \\
			%			\hline
			\Xhline{2.5\arrayrulewidth}
	\end{tabular}}
	\label{tab:quan1}
\end{table*}

%%%%%%%%%%%%%%%%%%%%%%%%%%%%%%%%%%%%%%%%% Experiments %%%%%%%%%%%%%%%%%%%%%%%%%%%%%%%%%%%%%%%%%%%%%%%
\section{Experiments}

In this section, we first introduce the benchmark datasets and their corresponding metrics, followed by the implementation details of our approach. 
We then compare our method with other MLLMs and present both quantitative and qualitative results. 
Finally, we conduct ablation experiments on some experimental settings and discuss its limitation.

\subsection{Evaluation Datasets and Metrics}
To comprehensively evaluate the effectiveness of our method, we conducted a wide range of experiments in various tasks, including image captioning, general VQA, scene text-centric VQA, key information extraction, and document-oriented VQA.

\begin{table}
	\centering
	\setlength\tabcolsep{1pt}
	\small
	\caption{Quantitative comparison with previous multimodal large language models (MLLMs) on document-oriented VQA. The best and the second results are highlighted in \textbf{bold} and \underline{underlined}, respectively.}
	\resizebox{0.98\linewidth}{!}{%
		\begin{tabular}{lcccccccc}
			\Xhline{2.5\arrayrulewidth}                                                  
			& DocVQA & ChartQA & InfoVQA & KLC & WTQ \\
			\hline
            BLIP2-OPT-6.7B~\cite{li2023blip} & 0.82 & 7.44 & 8.82 & - & \underline{17.82} \\
            LLaVAR~\cite{zhang2023llavar} & 6.73 & 8.00 & 12.25 & - & - \\
            UniDoc~\cite{feng2023unidoc} & 6.47 & 10.48 & 13.75 & - & -  \\
            mPLUG-Owl2~\cite{ye2023mplug2} & \underline{16.55} & \underline{18.72} & 14.28 & \underline{5.07} & -  \\
            LLaVA-1.5~\cite{liu2023improved} & 12.53 & 14.48 & \underline{15.74} & 2.59 & 11.54\\
            \textbf{AdaptVision}  & \textbf{48.61} & \textbf{30.96} & \textbf{21.21} & \textbf{32.74} & \textbf{24.80} \\
			%			\hline
			\Xhline{2.5\arrayrulewidth}
	\end{tabular}}
	\label{tab:quan2}
\end{table}

\begin{figure*}[tbp]
	\begin{center}
		\includegraphics[width=0.98\linewidth]{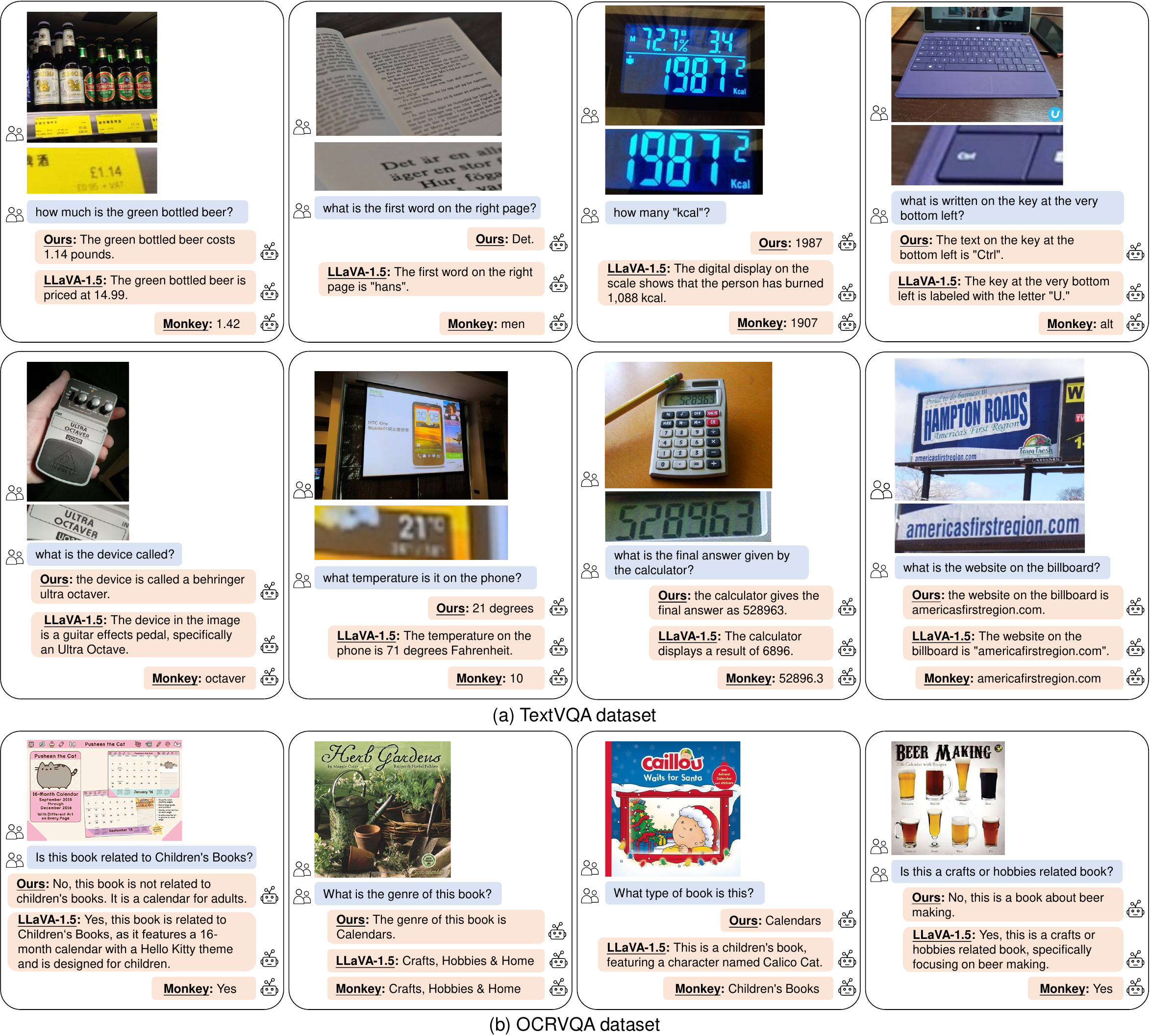}
	\end{center}
	\vspace{-0.1in}
	\caption{Visual comparison results of our AdaptVsion method with LLaVA-1.5~\cite{liu2023improved} and Monkey~\cite{li2023monkey} on TextVQA~\cite{singh2019towards} and OCRVQA~\cite{mishra2019ocr} datasets.}
	\label{fig:compare1}
\end{figure*}

\begin{figure*}[tbp]
	\begin{center}
		\includegraphics[width=0.98\linewidth]{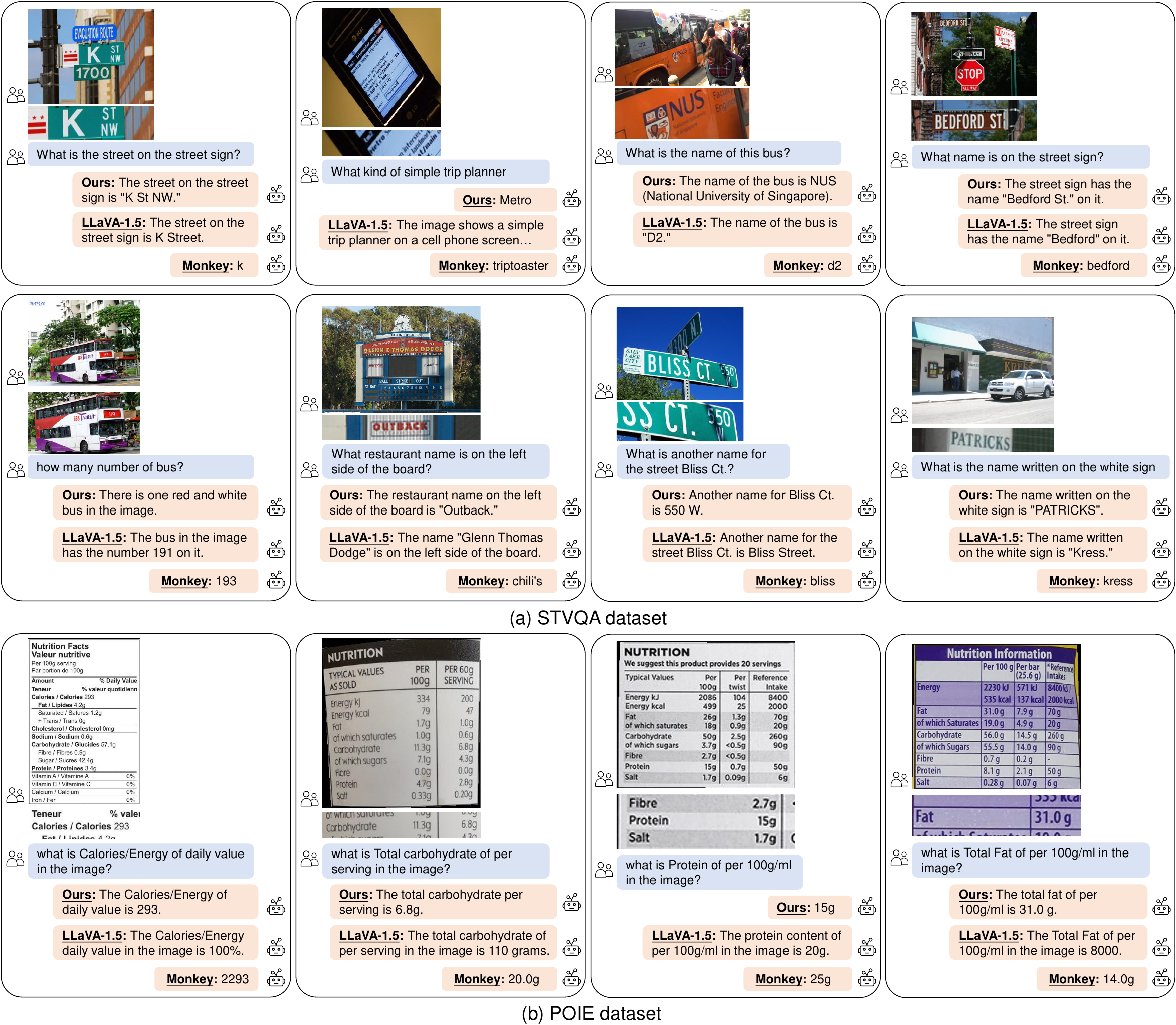}
	\end{center}
	\vspace{-0.1in}
	\caption{Visual comparison results of our AdaptVision method with LLaVA-1.5~\cite{liu2023improved} and Monkey~\cite{li2023monkey} on STVQA~\cite{biten2019icdar} and POIE~\cite{kuang2023visual} datasets.}
	\label{fig:compare2}
\end{figure*}

\begin{figure*}[tbp]
	\begin{center}
		\includegraphics[width=0.98\linewidth]{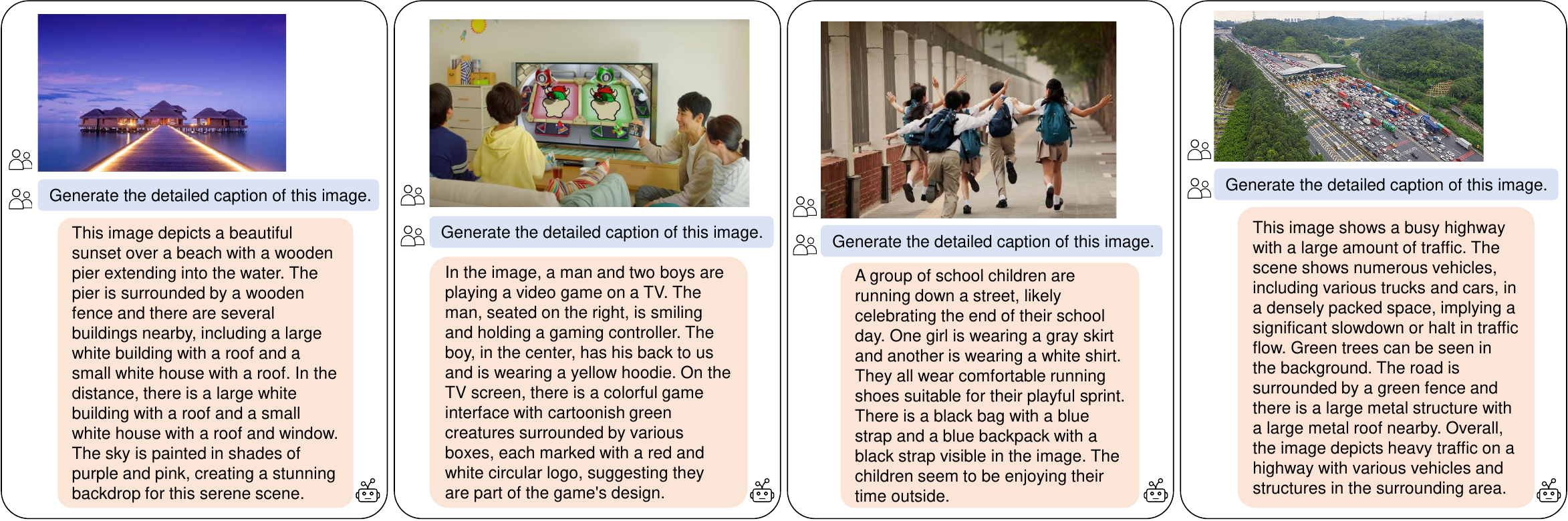}
	\end{center}
	\vspace{-0.1in}
	\caption{Visualization results of our AdaptVision method on image caption.}
	\label{fig:qua1}
\end{figure*}

For the image captioning task, we utilize the FLickr30K~\cite{young2014image} and TextCaps~\cite{sidorov2020textcaps} datasets, with CIDEr as the evaluation metric. 
For general VQA, our datasets include the VQAv2~\cite{goyal2017making}, OKVQA~\cite{marino2019ok}, GQA~\cite{hudson2019gqa}, ScienceQA~\cite{lu2022learn}, and VizWiz~\cite{gurari2018vizwiz} datasets. 
For scene text-centric VQA, we use STVQA~\cite{biten2019icdar}, OCRVQA~\cite{mishra2019ocr}, AI2D~\cite{kembhavi2016diagram}, and TextVQA~\cite{singh2019towards} datasets.
For document-oriented VQA, we utilize DocVQA~\cite{mathew2021docvqa}, ChartQA~\cite{masry2022chartqa}, InfoVQA~\cite{mathew2022infographicvqa}, KLC~\cite{stanislawek2021kleister}, and WTQ~\cite{pasupat2015compositional} datasets.
The evaluation metric for VQAv2, OKVQA and VizWiz is the VQA Score, while the metric for others is accuracy.
For the key information extraction task, we utilize FUNSD~\cite{jaume2019funsd}, SROIE~\cite{huang2019icdar2019}, and POIE~\cite{kuang2023visual} datasets, also with precision as an evaluation metric.

\begin{figure*}[tbp]
	\begin{center}
		\includegraphics[width=0.98\linewidth]{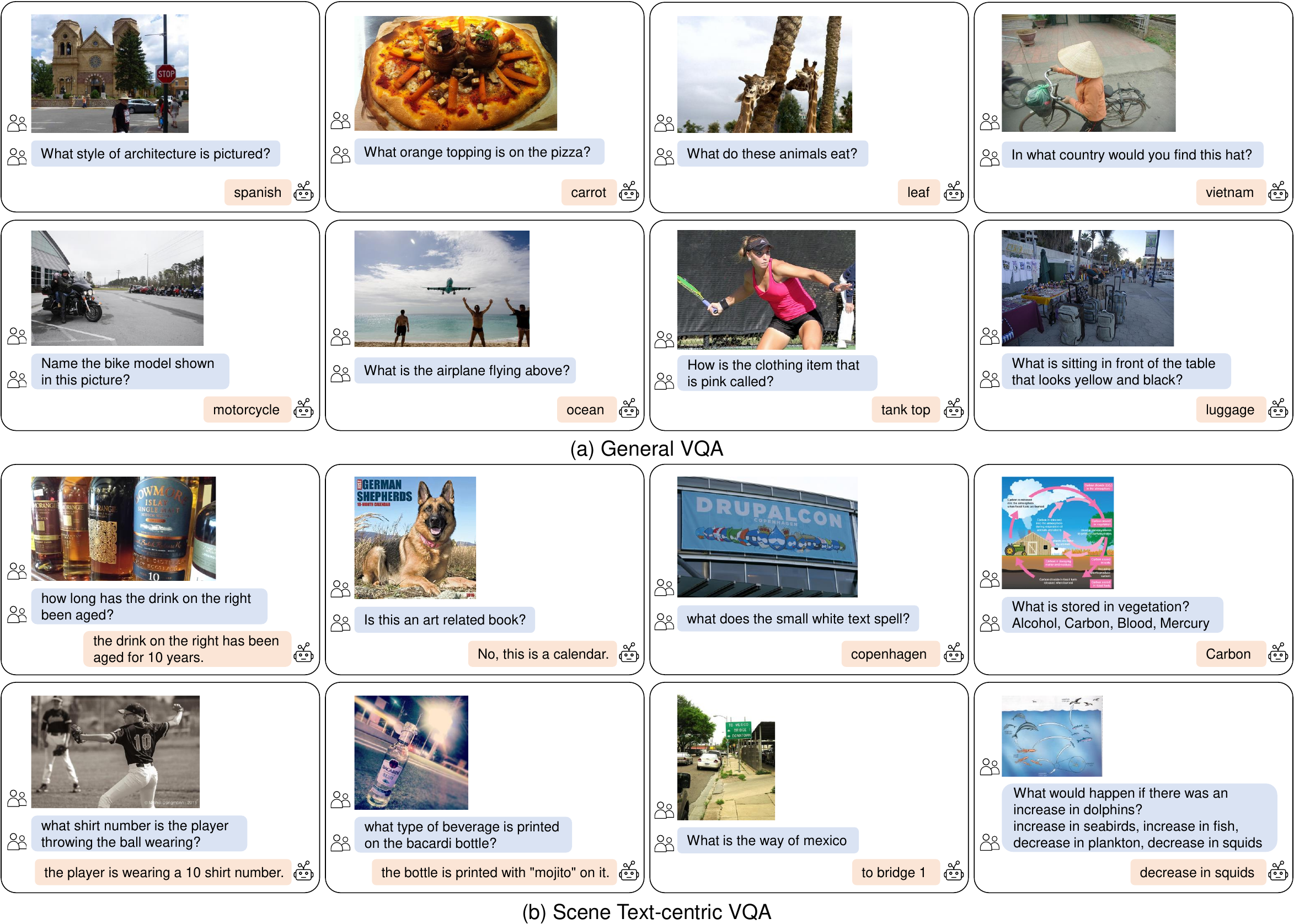}
	\end{center}
	\vspace{-0.1in}
	\caption{More visualization results of our AdaptVision method on general VQA and scene text-centric VQA tasks.}
	\label{fig:qua2}
\end{figure*}

\subsection{Implementation Details}
Our experiments were carried out on a Linux platform equipped with eight A100 GPUs.
The vision encoder of the model is CLIP-ViT-L-336px~\cite{radford2021learning}, incorporating a global branch and a dynamic local branch, each featuring a distinct CLIP encoder.
The projection layer consists of two simple linear layers, and the Large Language Model (LLM) employed is Vicuna-7B~\cite{chiang2023vicuna}, with a maximum sequence length set to 4096.
During the pre-training phase, we only fine-tuned the projector, setting the learning rate to 2e-3 and the batch size to 128.
For the fine-tuning phase, to overcome the CLIP's~\cite{radford2021learning} limitations in handling text and detailed scenarios due to its training on natural scenes, we chose to fine-tune the entire framework, including the vision encoder, projector, and LLM.
We set the learning rate to 2e-5 and the batch size to 32.
We used the AdamW~\cite{loshchilov2017decoupled} optimizer for updates and the cosine annealing scheduler for learning rate adjustment, completing a training epoch in each phase.

\subsection{Main Results}
We present the quantitative results of AdaptVision across various benchmarks, supplemented by qualitative results of the method as detailed below.

\textbf{Quantitative results.}
We perform a comparative analysis of our method against recent developments in multimodal large language models (MLLM) and present the results in Table~\ref{tab:quan1}.
We achieve the state-of-the-art results in eight out of fourteen benchmarks on four tasks and demonstrate strong competitiveness in the remaining six datasets.
Specifically, for image caption, our approach exhibits the best results on Flickr30K~\cite{young2014image} and competitive performance on the TextCaps~\cite{sidorov2020textcaps} benchmarks. 
Regarding general VQA, it maintains competitive performance in the OKVQA~\cite{marino2019ok} dataset, and achieves state-of-the-art results on VQAv2~\cite{goyal2017making}, GQA~\cite{hudson2019gqa}, scienceQA~\cite{lu2022learn} and VizWiz~\cite{gurari2018vizwiz} datasets. 
In terms of scene text-centric VQA, our approach achieves the best performance on the OCRVQA~\cite{mishra2019ocr} and AI2D~\cite{kembhavi2016diagram} datasets, specifically surpassing the Monkey~\cite{li2023monkey} by margins of 0.91 and 3.99 percentage points, respectively. 
For key information extraction, while our approach exhibits good results on the three benchmarks~\cite{jaume2019funsd,kuang2023visual,huang2019icdar2019}.
Furthermore, we also engage in document-oriented VQA challenges, as illustrated in Table~\ref{tab:quan2}, thus demonstrating the robustness of our method.

\begin{figure*}[tbp]
	\begin{center}
		\includegraphics[width=0.98\linewidth]{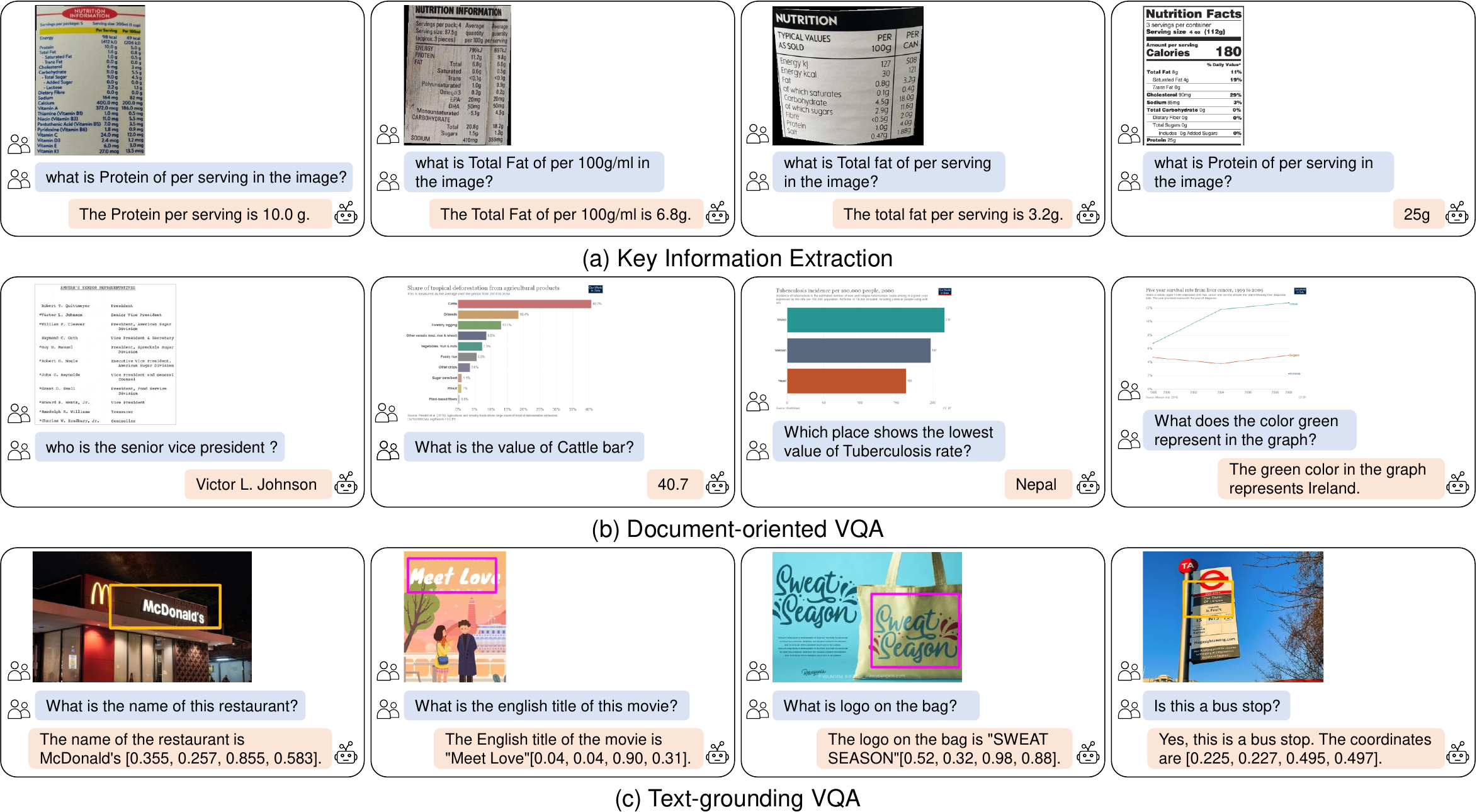}
	\end{center}
	\vspace{-0.1in}
	\caption{More visualization results of our AdaptVision method on key information extraction, document-oriented VQA, and text-grounding VQA tasks.}
	\label{fig:qua3}
\end{figure*}

\textbf{Qualitative results.}
We compare our method with LLaVA-1.5~\cite{liu2023improved} and Monkey~\cite{li2023monkey} on the TextVQA~\cite{singh2019towards}, OCRVQA~\cite{mishra2019ocr}, STVQA~\cite{biten2019icdar}, and POIE~\cite{kuang2023visual} datasets, with the visualization results shown in Figure~\ref{fig:compare1} and Figure~\ref{fig:compare2}. 
We find that our method can capture the fine-grained textual information in Scene Text-centric VQA scenarios well, demonstrating a better understanding of natural scenes containing text.
Moreover, We further provide more qualitative results across several tasks to validate the effectiveness of our method.
As shown in Figure~\ref{fig:qua1}, we showcase qualitative results for image caption, where our model demonstrates its proficiency in precisely identifying objects within images and generating detailed descriptions.
Figure~\ref{fig:qua2} offers examples for both general VQA and scene text-centric VQA tasks, indicating that our method is capable of accurately detecting content within images and generating appropriate responses to the given questions.
Figure~\ref{fig:qua3} presents instances of more complex tasks like key information extraction, document-oriented VQA, and text-grounding VQA, where our model also delivers accurate answers.
For text-grounding VQA, we randomly select images from the internet to the model along with queries.
The model not only provides answers but also elucidates its reasoning with bounding boxes, thereby enhancing the interpretability of the model.

\subsection{Ablation Study}
We conduct ablation studies on specific configurations to further validate the efficacy of our experimental setup. 
All experiments are performed on VQAv2~\cite{goyal2017making}, GQA~\cite{hudson2019gqa}, STVQA~\cite{biten2019icdar}, OCRVQA~\cite{mishra2019ocr}, AI2D~\cite{kembhavi2016diagram}, DocVQA~\cite{mathew2021docvqa}, and POIE~\cite{kuang2023visual} datasets.

% \begin{table}
% 	\centering
% 	\setlength\tabcolsep{1pt}
% 	\small
% 	\caption{Ablation study on predefined grids. In this paper, we use the $3\times 3$ grids in our experiments. The best results are highlighted in \textbf{bold}.}
% 	\resizebox{0.98\linewidth}{!}{%
% 		\begin{tabular}{cccccccc}
% 			\Xhline{2.5\arrayrulewidth}                                                  
% 			\multirow{3}{*}{Config} & \multicolumn{7}{c}{Benchmark} \\
%             % \cmidrule(lr){1-2} \cmidrule(lr){3-9}
%              \cmidrule(lr){2-8}
%                           & VQAv2 & GQA & STVQA & OCRVQA & AI2D & DocVQA & POIE \\
%             \hline
%             w/o local branch & 71.82 & 57.47 & 46.80 & 62.96 & 63.02 & 24.38 & 26.01 \\
%             w local branch   & \textbf{79.50} & \textbf{59.37} & \textbf{49.26} & \textbf{64.36} & \textbf{66.65}   & \textbf{48.61} & \textbf{36.67} \\
% 			\hline
% 			\Xhline{2.5\arrayrulewidth}
% 	\end{tabular}}
% 	\label{tab:abl_local}
% \end{table}

\begin{table}
	\centering
	\setlength\tabcolsep{1pt}
	\small
	\caption{Ablation study on predefined grids. In this paper, we use the $3\times 3$ grids in our experiments. The best results are highlighted in \textbf{bold}.}
	\resizebox{0.98\linewidth}{!}{%
		\begin{tabular}{cccccccc}
			\Xhline{2.5\arrayrulewidth}                                                  
			\multirow{3}{*}{Grid} & \multicolumn{7}{c}{Benchmark} \\
            % \cmidrule(lr){1-2} \cmidrule(lr){3-9}
             \cmidrule(lr){2-8}
                          & VQAv2 & GQA & STVQA & OCRVQA & AI2D & DocVQA & POIE \\
            \hline
            $2\times 2$ & 71.82 & 57.47 & 46.80 & 62.96 & 63.02 & 41.38 & 26.01 \\
            $3\times 3$ & \textbf{81.50} & \textbf{62.37} & \textbf{49.26} & \textbf{64.36} & \textbf{66.65} & \textbf{48.61} & \textbf{36.67} \\
			\hline
			\Xhline{2.5\arrayrulewidth}
	\end{tabular}}
	\label{tab:abl1}
\end{table}

% \textbf{Ablation study on the dynamic local branch.}
% Table~\ref{tab:abl_local}

\textbf{Ablation study on the pre-defined grids.}
To determine the impact of diversity in resolution and aspect ratio on the model's overall performance, we initially modified the predefined grids from $3\times 3$ to $2\times 2$ and executed experiments to observe the resultant effects. 
Table~\ref{tab:abl1} presents the quantitative results of both configurations. revealing that the method employing $3\times3$ grids achieved superior results in all benchmarks.
The results suggest that increasing resolution and diversifying aspect ratios can more effectively process local image information across various types while also mitigating the distortions introduced by image resizing.

% These results suggest that enhancing resolution and diversifying aspect ratios more effectively processes local image information across various types, simultaneously reducing distortions caused by image resizing.

\textbf{Ablation study on fine-tuning vision encoder.}
As previously mentioned, OpenAI's CLIP~\cite{radford2021learning}, trained on a wide range of natural scenes, often finds it challenging to accurately comprehend the fine-grained textual and detailed scene information. 
Consequently, during the training process, we fine-tuned the vision encoders in both the global and dynamic local branches to better adapt to diverse scenes. 
As shown in Table~\ref{tab:abl2}, we present the results of both fine-tuning the vision encoder and not fine-tuning it. 
The results intuitively reveal that fine-tuning the vision encoder enhances the capability of the CLIP to adapt to a variety of visual tasks more effectively, not merely confined to the understanding of natural objects, thereby yielding superior results.

\begin{table}
	\centering
	\setlength\tabcolsep{1pt}
	\small
	\caption{Ablation study on fine-tuning vision encoder. The best results are highlighted in \textbf{bold}.}
	\resizebox{0.98\linewidth}{!}{%
		\begin{tabular}{cccccccc}
			\Xhline{2.5\arrayrulewidth}                                                  
			\multirow{3}{*}{Config} & \multicolumn{7}{c}{Benchmark} \\
            % \cmidrule(lr){1-2} \cmidrule(lr){3-9}
             \cmidrule(lr){2-8}
                          & VQAv2 & GQA & STVQA & OCRVQA & AI2D & DocVQA & POIE \\
            \hline
            w/o finetune              & 73.51 & 58.45 & 47.90 & 56.08 & 61.07 & 35.69 & 31.04  \\
            w/ finetune & \textbf{81.50} & \textbf{62.37} & \textbf{49.26} & \textbf{64.36} & \textbf{66.65} & \textbf{48.61} & \textbf{36.67} \\
			\hline
			\Xhline{2.5\arrayrulewidth}
	\end{tabular}}
	\label{tab:abl2}
\end{table}

\subsection{Limitation}
% In this paper, we present AdaptVision for versatile scene understanding and demonstrate its impressive effectiveness. 
% However, as illustrated in Table~\ref{tab:quan2}, our method exhibits suboptimal overall performance for document-oriented VQA tasks. 
% Our analysis suggests that downsampling at the second linear layer, intended to reduce the number of visual tokens fed into the LLM, inadvertently leads to a loss of textual details.
% This process leads to a lack of coherence in the information encapsulated within the visual tokens, which adversely affects performance in text-dense scenarios. 
% In the future, we plan to refine our method by developing techniques for extracting key visual tokens from input image features, thereby preserving essential details.

% As shown in Table~\ref{tab:quan2}, our method exhibits suboptimal performance for document-oriented VQA. 
% We analyze that downsampling at the second linear layer, inadvertently leads to a loss of textual details and affects performance in text-dense scenarios. 
% In the future, we plan to refine our method by developing techniques to extract key visual tokens, thereby preserving essential details.

As shown in Table~\ref{tab:quan2}, our method exhibits suboptimal performance for document-oriented VQA. 
We observe that documents with resolutions above $1500\times 2000$ and InfoVQA~\cite{mathew2022infographicvqa} images have high resolution and large aspect ratios, such as $800\times 4500$.
Our method can handle a maximum resolution of $1008\times 1008$, which performs well in scene text-centric VQA but may not be sufficient for high-resolution, large aspect ratio fine-grained document images.
Specifically, our method might resize such images to $336\times 1008$, which fails to meet the resolution and aspect ratio requirements.
Consequently, our method may not achieve satisfactory results in such situations. 
To mitigate this issue, we plan to introduce a detection module in future work to identify dense text regions and specifically enlarge those areas, thereby improving the model's performance.

\section{Conclusion}
In this paper, we present AdaptVision, an approach designed to understand versatile scenes. 
The essence of our method is adaptively controlling the quantity of visual tokens input into the Large Language Model (LLM) according to the input image's size.
This method mitigates distortion effects that arise from resizing images to a uniform resolution and dynamically optimizing the visual tokens input to the LLMs.
Our approach has been empirically validated through a wide range of benchmarks, demonstrating the effectiveness of our method.

% \section*{Acknowledgments}
% This should be a simple paragraph before the References to thank those individuals and institutions who have supported your work on this article.

{\small
	\bibliographystyle{IEEEtran}
	\bibliography{egbib}
}

\vfill

\end{document}